\documentclass[runningheads]{llncs}

\usepackage[T1]{fontenc}

\usepackage{graphicx}

\begin{document}

\title{Data Synthesis and Parameter-Efficient Fine-Tuning for Low-Resource NMT: A Case Study on Q'eqchi' Mayan}

\titlerunning{Data Synthesis and PEFT for Q'eqchi' NMT}

\author{Alexander Chulzhanov\inst{1,2}\orcidID{0009-0005-5481-8179} \and
Soeren Eberhardt\inst{3}\orcidID{0009-0005-7643-9743} \and
Arjun Mukherjee\inst{1}\orcidID{0000-0002-8896-604X}}

\authorrunning{A. Chulzhanov et al.}

\institute{University of Houston, Houston TX 77004, USA
\email{achulzha@cougarnet.uh.edu} \and
MasterWord Services, Inc., Houston TX 77079, USA \and
University of Washington, Seattle WA 98195, USA}

\maketitle

\let\thefootnote\relax\footnotetext{This version of the contribution has been accepted for publication, after peer review but is not the Version of Record and does not reflect post-acceptance improvements, or any corrections. The Version of Record is available online at: http://dx.doi.org/[insert DOI]. Use of this Accepted Version is subject to the publisher’s Accepted Manuscript terms of use https://www.springernature.com/gp/open-research/policies/accepted-manuscript-terms}

\begin{abstract}
Neural machine translation for digitally low-resource Indigenous languages is often hindered by extreme data scarcity, prompting reliance on extractive web-scraping. To ensure data sovereignty, this study introduces a data synthesis methodology to bootstrap NMT models without scraping target-language parallel text. Focusing on Q'eqchi' Mayan, we transformed community-sourced dictionaries into a massive synthetic corpus, utilizing Parameter-Efficient Fine-Tuning (PEFT) via LoRA adapters on an mT5-base model.

In-domain evaluation demonstrates high structural acquisition (BLEU 42.02), proving that synthetic constraints effectively teach complex agglutinative morphology and VOS word order. However, evaluation against an organic glossary reveals a structural-semantic gap (BLEU 0.59), where the model maintains grammatical integrity but lacks the lexical grounding of natural language. The model exhibits overfitting to the constrained structural variance of the synthetic templates; despite high semantic entropy in the pipeline, it struggles with the syntactic fluidity of natural language, forcing organic inputs into rigid learned patterns. Furthermore, an ablation study utilizing a Multi-Task Learning architecture resulted in negative transfer, suggesting that auxiliary tasks competed for limited parameter capacity within the LoRA adapters, causing over-optimization for synthetic markers at the expense of organic flexibility. Ultimately, we establish that synthetic bootstrapping is a highly effective structural primer, but requires authentic data for semantic refinement via Curriculum Learning.

\end{abstract}

\section{Introduction}

Neural Machine Translation (NMT) relies heavily on massive web-scraped datasets, creating a fundamental ``cold start'' problem for digitally low-resource Indigenous languages. Opportunistic web-scraping frequently violates data sovereignty and introduces severe domain skew, as public corpora are overwhelmingly dominated by archaic religious texts. Motivated by a strict ``Zero Target-Language Scrape'' policy, this research investigates how to ethically bootstrap a high-fidelity NMT system. While we leverage a foundational model pre-trained on high-resource languages, we ensure the Indigenous target data is sourced exclusively from contemporary community dictionaries without relying on web-scraped parallel text.

While synthetic data generation is a common augmentation technique, its efficacy as the sole foundational training mechanism for morphologically rich languages remains under-explored. Standard open-source NMT baselines do not currently exist for Q'eqchi', and anecdotal evaluations of commercial translation systems by fluent speakers reveal severe degradation and archaic bias. Therefore, a primary objective of this study is to establish a reproducible, zero-scrape baseline. Specifically, it is unclear if rule-based templates can successfully teach a pre-trained transformer procedural syntax without the semantic grounding of human-translated corpora. In this paper, we evaluate the structural and semantic boundaries of synthetic bootstrapping. We developed a distribution-aware, template-based generation pipeline to transform static Q'eqchi' dictionaries into a large-scale synthetic corpus. We fine-tuned an mT5-base model using Parameter-Efficient Fine-Tuning (LoRA) and focal loss optimization to map Q'eqchi' syntax to the model's pre-trained English and Spanish latent spaces. Additionally, an ablation study using a Multi-Task Learning (MTL) architecture tested whether auxiliary part-of-speech tagging could anchor semantics during synthetic training.

Our evaluation reveals a distinct dichotomy in model acquisition. The model successfully acquired complex agglutinative morphology and Verb-Object-Subject (VOS) word order, proving that synthetic constraints solve the syntactic cold start problem. However, evaluation against an out-of-domain organic glossary resulted in semantic collapse, where the model reverse-engineered templates but hallucinated vocabulary. Crucially, the MTL ablation study resulted in negative transfer, demonstrating that auxiliary tasks compete with primary translation objectives and cannot compensate for ungrounded training data. 

Ultimately, we demonstrate that data synthesis is highly effective for establishing a grammatical foundation, but requires organic data for lexical resolution. The main contributions of this work are: (1) A distribution-aware synthesis pipeline designed to ethically bootstrap NMT without scraping target-language text. (2) Empirical proof that parameter-efficient fine-tuning on synthetic data successfully teaches procedural grammar and VOS word order. (3) An ablation study demonstrating that MTL architectures cause negative transfer in LoRA adapters under purely synthetic constraints. (4) A proposed Curriculum Learning framework utilizing synthetic data strictly as a structural primer prior to authentic semantic refinement.

\section{Related Work}

\textbf{Low-Resource Translation and Data Scarcity.}
The transition to massively multilingual models, such as NLLB-200 \cite{nllb2024} and mT5 \cite{xue2020mt5}, has established new baselines for machine translation. However, these models heavily rely on massive, web-scraped corpora. For Indigenous languages, this reliance introduces severe domain skew and often violates data sovereignty. Recent studies evaluating multilingual models in low-resource settings \cite{scalvini2025} validate that while foundational models offer robust cross-lingual transfer, they require targeted, domain-specific fine-tuning to recover local nuances and avoid archaic biases. 

\textbf{Synthetic Data Generation.}
To address extreme data scarcity, recent literature increasingly turns to synthetic data generation. Contemporary approaches often utilize high-capacity LLMs to hallucinate parallel corpora, such as the SynOPUS framework \cite{synopus2025}, which demonstrates that even noisy synthetic alignments can improve translation performance. In contrast to these probabilistic LLM approaches, our work investigates the efficacy of deterministic, rule-based synthetic generation to build an exact, grammatically perfect scaffolding without the risk of initial semantic hallucination.

\textbf{Syntactic Scaffolding and Multi-Task Learning.}
Bridging the morphological gap in low-resource settings frequently involves utilizing syntactic markers as a universal cross-lingual signal. Research has demonstrated that utilizing Part-of-Speech (POS) sequences as a structural backbone \cite{senticvec2024} and integrating rule-encoded linguistic constraints directly into the training objective \cite{r2t2025} significantly reduces the need for extensive lexical training data. We build upon this concept by evaluating whether MTL principles hold true when constrained entirely to synthetic adapters, or if the lack of organic semantic grounding causes architectural interference.

\section{Methodology: Synthetic Generation}

At the core of our methodology is a deterministic, rule-based synthetic data generator designed to produce massive-scale parallel corpora. Unlike stochastic LLM generation which can be prone to hallucination, our system operates on strict linguistic constraints to ensure grammatical fidelity across the target triplet: Q’eqchi’, English, and Spanish. This engine transforms static dictionary entries into dynamic, aligned sentence pairs that serve as the ground truth for training the NMT model. To ensure reproducibility, the complete pipeline has been made available via GitHub and Hugging Face repositories.\footnote{Code and synthetic datasets are available at: \url{https://github.com/achulzhanov/mayan-mt5}}

\begin{figure}[htbp]
    \centering
    \includegraphics[width=\linewidth]{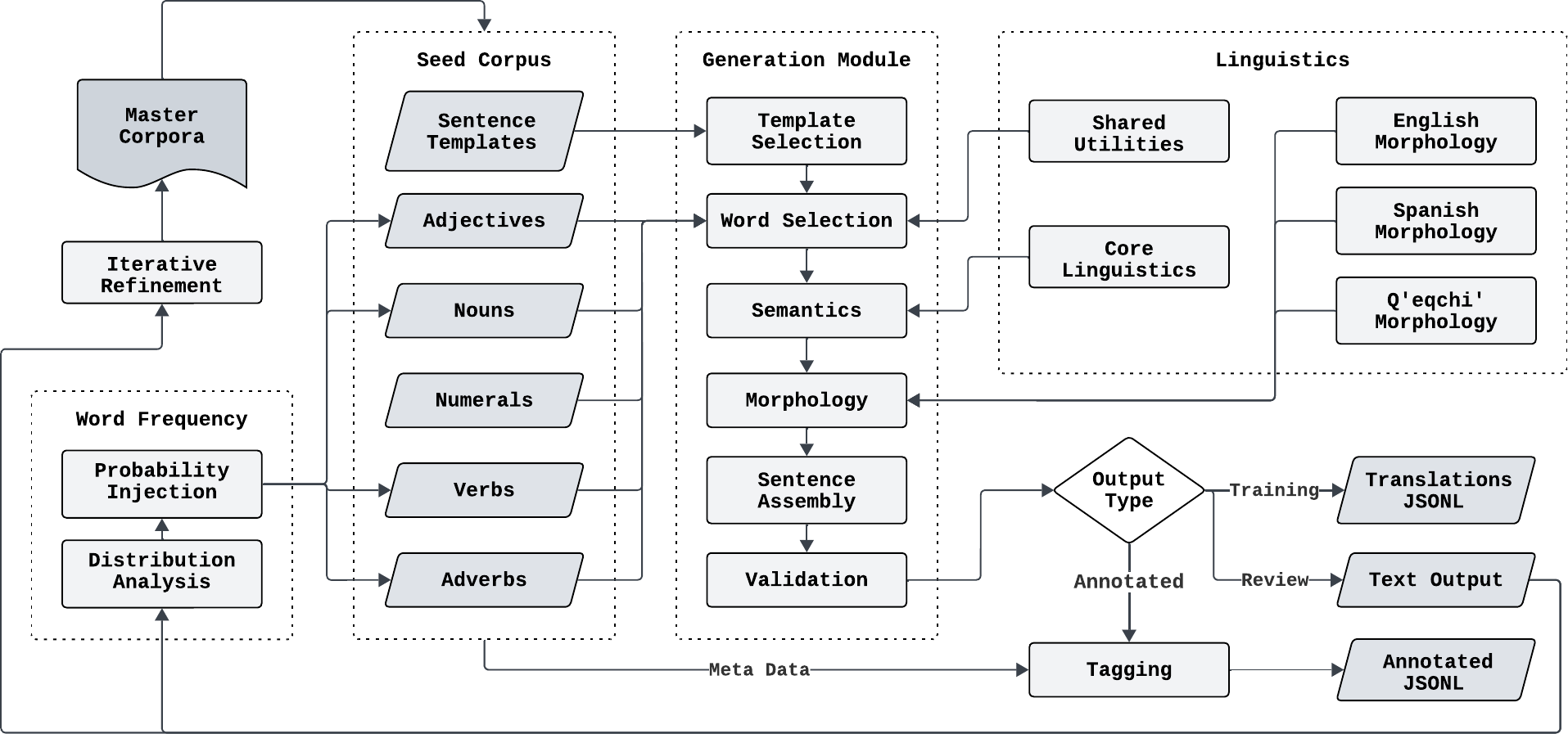}
    \caption{High-level overview of the synthetic sentence generator program written in Python.}
    \label{fig:generator_overview}
\end{figure}

\subsection{Structured Seed Data and Linguistic Modules}

Our generator, as illustrated in Fig.~\ref{fig:generator_overview}, utilizes structured lexical databases (CSVs) containing semantic alignments across Q'eqchi', English, and Spanish, enriched with morphological metadata (e.g., transitivity, possession classes). These terms dynamically populate placeholders within a registry of syntactic templates. This decoupling of lexical content from structural patterns enables a single template to generate thousands of grammatically correct sentence permutations while preserving native word order across all three languages.

To manage multi-language morphology efficiently, the generator employs a modular strategy pattern. Language-specific modules handle all inflection logic, such as verb conjugation, noun pluralization, and structural agreement, independently of the core generation engine. This ensures the pipeline remains language-agnostic, allowing seamless adaptation to other indigenous languages.

\subsection{Word Frequency Distribution and Zipf's Law}

Standard synthetic generation relies on uniform random sampling, which fails to reflect the natural frequency distributions of language governed by Zipf's Law. To construct realistic training distributions without existing Q'eqchi' corpora, we implemented a max-proxy weighting system. The generator estimates a Q'eqchi' term's frequency by utilizing the maximum frequency of its English or Spanish equivalents, ensuring cultural and semantic salience.

To prevent high-frequency terms from monopolizing the dataset, we apply distribution smoothing. This redistributes probability mass from the top percentile of words to the ``long tail'' of the vocabulary, guaranteeing that rare terms still achieve sufficient representation for stable embedding space mapping.

\subsection{Gender Debiasing and Contextual Resolution}

In production NMT systems, the failure to accurately recover grammatical gender is frequently misattributed to inherent algorithmic bias, when it is often a symptom of underspecified source context. To prevent the inadvertent amplification of these errors, particularly in Spanish translations, we implemented explicit debiasing mechanisms and contextual ``hints'' within the template logic.

For ambiguous copular constructions where gender is not syntactically recoverable (e.g., English ``You are beautiful''), the generator enforces symmetric randomization of Spanish adjectival gender. Conversely, to provide the model with sufficient context to accurately resolve gender when appropriate, templates are systematically enriched with paired, gender-specific names and explicitly marked occupations. These deterministic constraints ensure training parity without relying on opaque, post-hoc statistical reweighting.

\subsection{Generation Loop}

The core generation engine operates as a constraint satisfaction system executed in five stages: (1) \textit{Template Selection} from the syntactic registry; (2) \textit{Constraint Solving} for semantic slot requirements; (3) \textit{Weighted Slot Filling} utilizing max-proxy probabilities; (4) \textit{Morphological Inflection} via isolated language modules; and (5) \textit{Validation} to ensure linguistic constraints are strictly met prior to serialization.

\subsection{Data Serialization and Task Homogenization}

Generated sentences are serialized into the JSON Lines (JSONL) format, ensuring direct compatibility with modern transformer training pipelines. For the standard syntactic bootstrapping phase, the data is structured as aligned source and target strings alongside language codes:

\begin{verbatim}
{"translation": {"src_lang": "eng", "tgt_lang": "kek", 
 "src_text": "Their sister didn't float.", 
 "tgt_text": "Moko xpamamnak ta li anab'."}}
\end{verbatim}

To support the Multi-Task Learning (MTL) ablation study, a dedicated tagging module intercepts morphological metadata during generation, outputting parallel structural datasets alongside standard text. The resulting schema appends keys for part-of-speech (\texttt{pos\_kek}) utilizing Universal Dependencies POS tags \cite{ud_pos}, and semantic tagging (\texttt{semantic\_kek}), capturing granular details like named entity recognition:

\begin{verbatim}
{"translation": {...,
 "src_text": "Aj Chisecat. Laa'at aj yoob'ahom kab'l.",
 "tgt_text": "You are from Chisec. You are an architect.",
 "pos_kek": "Aj (DET) Chisecat (PROPN) ... 
             yoob'ahom (NOUN) kab'l (NOUN) . (PUNCT)",
 "semantic_kek": "... Chisecat (B-LOCATION) ... 
                  yoob'ahom (B-OCCUPATION) kab'l (B-OCCUPATION)"}}
\end{verbatim}

Because mT5 is a generalized text-to-text model, it processes all inputs and outputs as continuous strings. During collation, JSON objects are parsed into text prompts using explicit task prefixes (e.g., ``translate English to Q'eqchi':''). By standardizing the input string structure, the model is forced to share its latent representations across all tasks.

\section{Experimental Setup}

\subsection{Model Selection and Tokenization}

We utilized Google’s mT5-base (Multilingual Text-to-Text Transfer Transformer) as our foundational model. Initial exploratory experiments focused exclusively on the translation-specific NLLB-200 architecture \cite{nllb2024}. To accommodate Q'eqchi' agglutinative morphology, we attempted to expand the NLLB tokenizer vocabulary with approximately 40 language-specific morphemes. However, initializing these new token embeddings introduced severe gradient instability, as the model struggled to simultaneously learn the Q'eqchi' structural rules and the unaligned embedding weights. 

To resolve this, we pivoted to generalized text-to-text architectures. We initially evaluated the standard T5 model before ultimately selecting mT5-base due to its significantly stronger Spanish pre-training and robust inter-language representations, simplifying the training objective from learning translation mechanics to mapping existing probability vectors to Q'eqchi' surface forms. Furthermore, we abandoned manual vocabulary expansion entirely; because Q'eqchi' utilizes a Latin-based orthography, mT5’s default SentencePiece tokenizer demonstrated sufficient zero-shot subword segmentation, completely eliminating the gradient instability associated with untrained token embeddings.

\subsection{Parameter-Efficient Fine-Tuning (LoRA)}

To mitigate the risk of catastrophic forgetting of pre-trained priors, we employed Low-Rank Adaptation (LoRA). We targeted both the attention modules (\texttt{q, v, k, o}) for syntactic routing and feed-forward networks (\texttt{wi\_0, wi\_1, wo}) to maximize influence over factual storage. We utilized a rank ($r$) of 32, alpha of 32, and 0.2 dropout, reducing trainable parameters to approximately 13.57M (2.3\% of the base model) for full local training. Optimization proceeded for 3 epochs (or until convergence) using AdamW, a \texttt{1e-4} learning rate, an effective batch size of 8, and a max gradient norm of 10.0.

\subsection{Data Preparation and MTL Architecture}

To maximize the utility of scarce resources, we adopted a curriculum learning strategy. Phase 1 (the scope of this study) utilizes purely synthetic data to teach procedural syntax, while Phase 2 (future work) will introduce authentic glossaries to resolve semantic gaps. 

For the baseline translation adapter, the generator produced approximately 900,000 training pairs. To support the MTL ablation study, empty annotation rows were explicitly filtered, resulting in an organic task imbalance of approximately 4.2:1:1 (Translation:POS:Semantic). The expanded MTL training corpus contained 1,324,960 records. To ensure fair evaluation, validation was strictly limited to a fixed subset of 1,000 translation-only samples, completely isolating the evaluation loop from the auxiliary tasks.

\subsection{Focal Loss Optimization and Task Weighting}

In standard language modeling, Cross-Entropy loss is utilized to match natural word frequency distributions. However, our synthetic fine-tuning task relies heavily on mapping high-frequency structural tokens (e.g., articles, pronouns) to existing latent vectors. Empirical tests resulted in gradient domination; the sheer volume of highly confident structural predictions drowned out the subtle adjustments required to learn complex Mayan morphology. 

To resolve this, we implemented focal loss. A modulating factor ($\gamma = 2.0$) suppresses the loss contribution of well-classified structural examples to near zero, forcing the optimizer to dedicate limited adapter capacity entirely to rare suffixes.

Furthermore, to accommodate the 4.2:1:1 data imbalance in the MTL architecture, focal loss was computed per example and multiplied by a designated task weight before batch averaging. We calibrated the task weights to 1.0 for translation, 0.4 for POS, and 0.8 for semantic tagging, ensuring the auxiliary tasks maintained sufficient gradient influence without destabilizing the primary objective.

\subsection{Training Dynamics and Convergence}

\begin{figure}[htb]
    \centering
    % First sub-figure (STL)
    \begin{minipage}{0.48\textwidth}
        \centering
        \includegraphics[width=\textwidth]{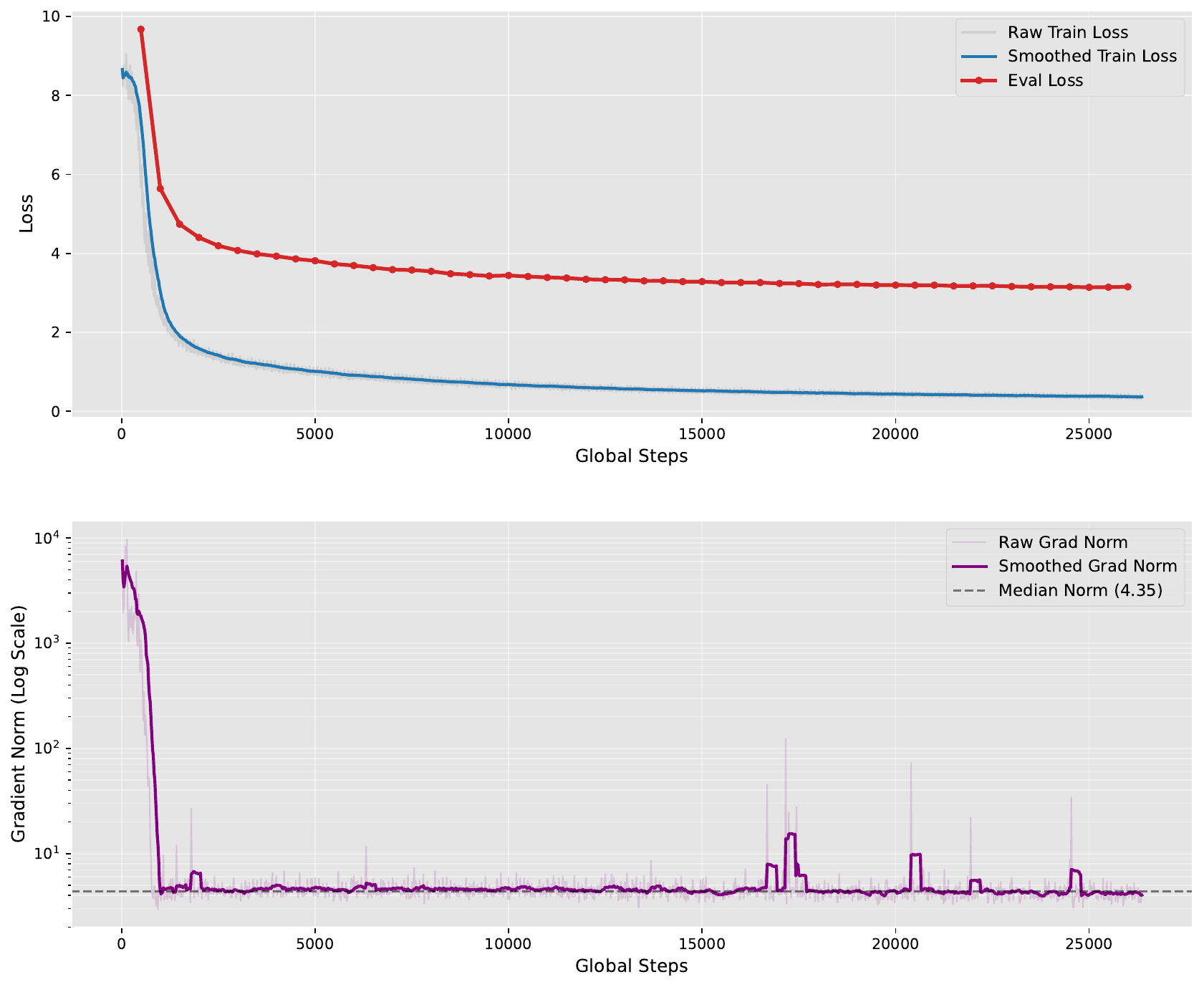}
        \caption{STL training dynamics showing rapid structural convergence.}
        \label{fig:stl_training}
    \end{minipage}\hfill
    % Second sub-figure (MTL)
    \begin{minipage}{0.48\textwidth}
        \centering
        \includegraphics[width=\textwidth]{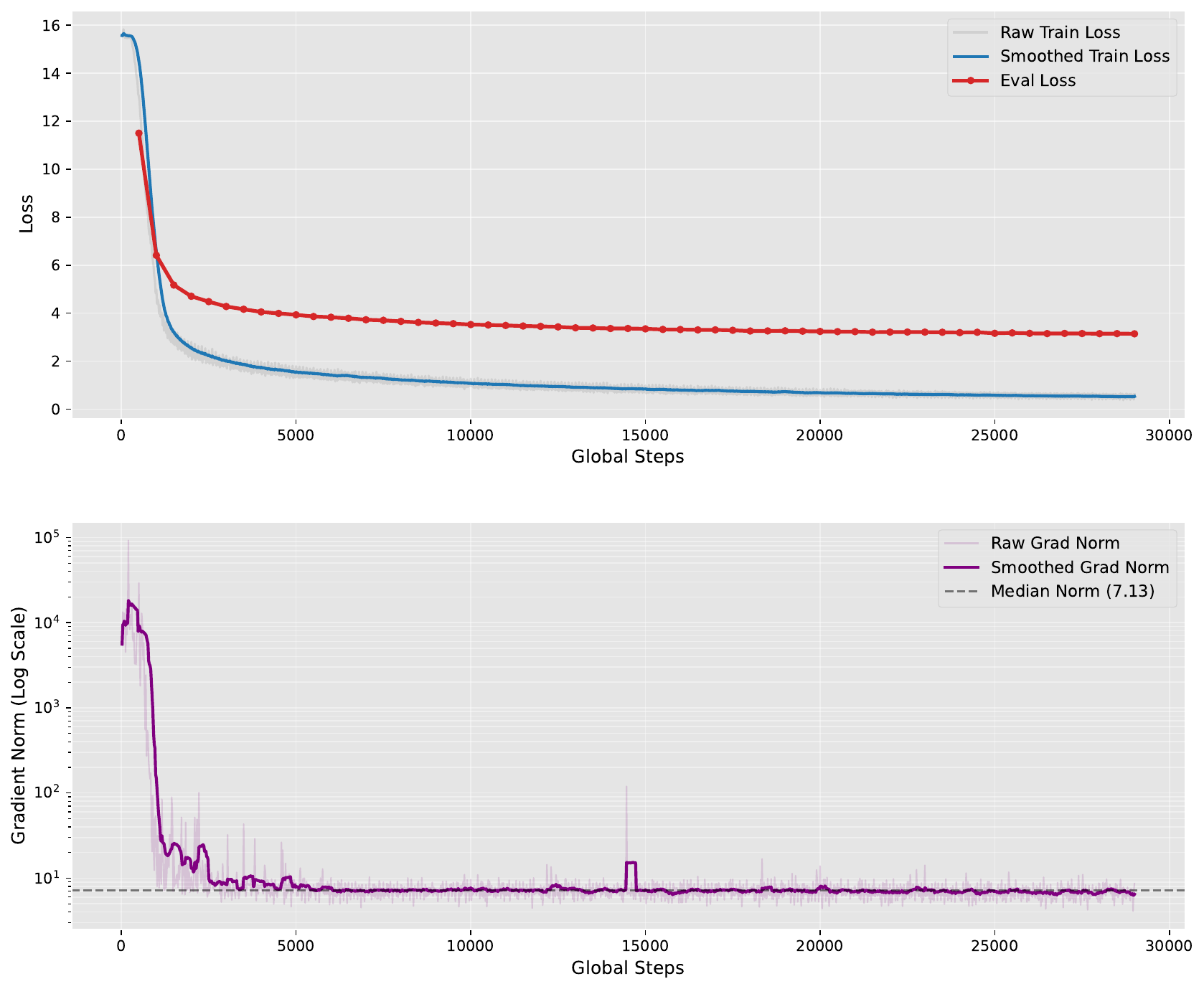}
        \caption{MTL training dynamics demonstrating elevated gradient noise.}
        \label{fig:mtl_training}
    \end{minipage}
\end{figure}

The training and evaluation loss curves across both architectures exhibit a distinct inversion characteristic of focal loss optimization. The respective optimization behaviors are plotted in Fig.~\ref{fig:stl_training} and Fig.~\ref{fig:mtl_training}. While training loss typically remains higher than evaluation loss due to LoRA dropout, our models demonstrate the opposite relationship. The STL training loss dropped to a floor of 0.4, and the MTL training loss settled near 0.5. Conversely, evaluation loss stabilized around 3.1. This inversion validates the focal loss function: as the model masters highly frequent structural templates, the modulating factor drives their loss to near zero. The evaluation loss remains higher because the validation set introduces unseen vocabulary combinations, representing the irreducible complexity of the morphological prediction task when the model cannot simply memorize specific template instances.

An analysis of gradient norms illustrates differing optimization stabilities. Both models experienced an initial shock phase during the first 2,000 steps as the adapter adjusted to the Q'eqchi' embedding space. Following this, the STL model settled into a quiet optimization plateau (median gradient norm 4.35). In contrast, the MTL model maintained a significantly higher median norm of 7.13. This elevated noise floor reflects the continuous gradient competition introduced by the auxiliary tasks. Consequently, the STL model achieved structural convergence significantly faster, suggesting the limited LoRA capacity struggles to simultaneously map complex morphology and auxiliary annotations without extending the optimization timeline.

Finally, while specific translation performance will be detailed in the subsequent section, we observed significant volatility in evaluation metrics during the training checkpoints. With a restricted validation set of 1,000 samples generated from shared templates, a minor shift in the probability distribution of a single token can alter the beam search path for hundreds of similar sentences simultaneously. Therefore, these metric fluctuations should be interpreted as artifacts of the synthetic validation methodology, where the model either perfectly matches a batch template or universally misses a shared morphological rule, rather than an indication of true model instability.

\section{Results and Evaluation}

\subsection{Evaluation Methodology}

To accurately measure structural acquisition and semantic retention, evaluation utilized 1,250 unique synthetic sentence pairs, expanded across all four translation directions to yield 5,000 evaluated pairs. The out-of-domain evaluation utilized an organic, human-translated glossary yielding 6,610 evaluated pairs. Translation quality was measured using BLEU, chrF, and Translation Edit Rate (TER).

Due to strict space constraints, metrics are reported as aggregates across all four translation directions. It should be noted that achieving these stable evaluations required extensive preliminary experimentation. Prior training attempts utilizing standard cross-entropy loss and uniform data distribution routinely resulted in severe gradient explosions and catastrophic weight corruption, ultimately necessitating the introduction of the focal loss and max-proxy weighting methodologies detailed in Section 4.

\begin{table}[htb]
\caption{Translation performance across architectures and evaluation domains. Upward arrows ($\uparrow$) indicate higher scores are better, while downward arrows ($\downarrow$) indicate lower scores are better.}\label{tab:translation_metrics}
\centering
\begin{tabular}{lcccccc}
\hline
\textbf{Model Architecture} & \multicolumn{3}{c}{\textbf{In-Domain (Synthetic)}} & \multicolumn{3}{c}{\textbf{Out-of-Domain (Organic)}} \\
\cline{2-4} \cline{5-7}
 & \textbf{BLEU $\uparrow$} & \textbf{chrF $\uparrow$} & \textbf{TER $\downarrow$} & \textbf{BLEU $\uparrow$} & \textbf{chrF $\uparrow$} & \textbf{TER $\downarrow$} \\
\hline
Baseline (Translation Only) & 42.02 & 59.69 & 41.53 & 0.59 & 12.51 & 94.40 \\
MTL (Trans + POS + Sem) & 46.97 & 63.27 & 37.15 & 0.48 & 12.26 & 94.14 \\
\hline
\end{tabular}
\end{table}

\subsection{In-Domain Acquisition and Prior Leverage}

As detailed in Table \ref{tab:translation_metrics}, the baseline model achieved a BLEU score of 42.02 on the in-domain test set, demonstrating that synthetic bootstrapping effectively maps procedural grammar to the mT5 latent space. Crucially, qualitative analysis revealed instances where the model's generation was stylistically and semantically flexible, rather than rigidly adhering to the synthetic ground truth. This indicates that the LoRA adapter successfully leverages the robust pre-trained priors of English and Spanish, allowing the model to map Q'eqchi' concepts to a broader latent semantic space rather than blindly memorizing exact string pairs.

\begin{table}[htb]
\caption{Example of in-domain prior leverage resulting in valid synonym generation.}\label{tab:indomain_example}
\centering
\begin{tabular}{p{0.35\linewidth} p{0.3\linewidth} p{0.25\linewidth}}
\hline
\textbf{Prompt (KEK to SPA)} & \textbf{Target (Synthetic)} & \textbf{Prediction (Model)} \\
\hline
Nab'ek laj Carlos. & Carlos anda. & Carlos camina. \\
\hline
\end{tabular}
\end{table}

For example, as shown in Table \ref{tab:indomain_example}, when translating the Q'eqchi' sentence \textit{``Nab'ek laj Carlos''} into Spanish, the synthetic generator produced the target \textit{``Carlos anda''} (English: ``Carlos goes/walks''). However, the model predicted \textit{``Carlos camina''} (English: ``Carlos walks''). Rather than strictly memorizing the exact synthetic vocabulary artifact, the model successfully mapped the Q'eqchi' verb \textit{nab'ek} to its underlying semantic representation, utilizing a perfectly valid Spanish synonym from its pre-trained priors to construct the output.

\subsection{Out-of-Domain Structural-Semantic Gap}

Evaluation against the organic glossary yielded a baseline BLEU score of 0.59. Rather than a simple performance collapse, this severe degradation highlights a distinct structural-semantic gap. The synthetic seed corpus was restricted to general domains, whereas the organic glossary contained specialized medical and legal terminology. Consequently, the model suffered systemic out-of-vocabulary errors. It effectively reverse-engineered the syntactic templates but hallucinated missing lexical tokens. 

\begin{table}[htb]
\caption{Out-of-domain evaluation examples demonstrating semantic hallucination due to template overfitting (top) and successful VOS structural acquisition (bottom).}\label{tab:outofdomain_examples}
\centering
\begin{tabular}{p{0.32\linewidth} p{0.3\linewidth} p{0.3\linewidth}}
\hline
\textbf{Prompt} & \textbf{Target (Organic)} & \textbf{Prediction (Model)} \\
\hline
\textit{(KEK to ENG)} Jun maak b'arwi' junaq poyanam naxk'e rib'... & The offense of willfully telling an untruth in a court... & She gives the snake to sixteen old ladies. \\
\hline
\textit{(ENG to KEK)} The hearing was scheduled in order to take pleadings from the defendant. & K'ojob'anb'il xq'ehil li ch'utam re xk'ulb'aleb' li junjunq xhuhil yaalalil li jitb'il. & Xto' li aj tz'iib'anel amaq' tenamit li aj raqol aatin. \\
\hline
\end{tabular}
\end{table}

For instance, as shown in the top row of Table \ref{tab:outofdomain_examples}, when presented with the complex Q'eqchi' legal definition for perjury, the model failed to translate out-of-vocabulary concepts. Instead, it forcefully mapped the unfamiliar input onto a rigid English template memorized during synthetic training. While it produced a grammatically perfect English sentence, it hallucinated the semantic vocabulary entirely from the generic seed dictionary.

However, as demonstrated in the bottom row of Table \ref{tab:outofdomain_examples}, this semantic gap simultaneously serves as a diagnostic indicator for structural acquisition. When translating an English SVO prompt into Q'eqchi', the model encountered out-of-vocabulary legal terms. Instead of failing to generate output or defaulting to English word order, the model applied the acquired Q'eqchi' VOS structure. It generated ``Xto' li aj tz'iib'anel amaq' tenamit li aj raqol aatin'' (literally: ``Lent the people's clerk the judge.''), accurately executing the Verb-Object-Subject morphology despite possessing the wrong semantic vocabulary. While comprehensive quantitative probing is required to make absolute claims about grammatical fluency, this qualitative example suggests strong structural entrenchment of target language grammar rules despite severe out-of-vocabulary constraints.

\subsection{MTL Ablation and Capacity Bottlenecks}

The Multi-Task Learning ablation study yielded an unexpected dichotomy. The MTL architecture outperformed the baseline in-domain (BLEU 46.97 vs 42.02) but suffered worse degradation out-of-domain (BLEU 0.48 vs 0.59). By forcing the adapter to predict semantic and morphological metadata, the model learned to explicitly memorize our generator's underlying mathematical patterns. This negative transfer suggests that auxiliary tasks competed for limited parameter capacity within the LoRA adapters (just 2.3\% of the base model), causing the model to over-optimize for synthetic markers at the expense of organic flexibility. While this negative transfer points to a capacity bottleneck, future work will evaluate whether scaling the LoRA rank ($r$) beyond 32 can alleviate this parameter pressure and mitigate architectural interference during joint optimization.

\subsection{Translation as Arithmetic: Base-Conversion Bottleneck}

While we did not conduct isolated quantitative experiments exclusively on arithmetic translation, our qualitative error analysis revealed a consistent base-conversion bottleneck. English and Spanish utilize a base-10 decimal system, whereas Q'eqchi' employs a base-20 vigesimal system. Standard sequence-to-sequence autoregressive transformers process numerical values as semantic text tokens rather than discrete, calculable mathematical entities. Because the foundational model's weights are fundamentally pre-trained on decimal string frequencies, mapping them to a vigesimal syntactic structure introduces systemic token-alignment conflicts that sequence prediction alone struggles to reconcile. Consequently, we propose that production systems should bypass neural sequence prediction for arithmetic entirely, relying instead on a deterministic, programmatic post-processing pipeline to intercept, convert, and format numerical values outside of the transformer block.

\section{Conclusion}
In this paper, we introduced a rule-based data synthesis methodology to ethically bootstrap a Q'eqchi' Mayan translation model without relying on extractive web scraping. By fine-tuning an mT5-base model exclusively on synthetic templates using LoRA and focal loss optimization, we successfully demonstrated that generative constraints can solve the syntactic cold start problem for low-resource languages. The baseline model achieved robust in-domain performance, proving its capacity to acquire complex agglutinative morphology and VOS word order while leveraging the strong pre-trained priors of English and Spanish. Crucially, Focal Loss enabled this structural acquisition; without it, high-frequency structural tokens would monopolize the limited adapter capacity, causing the model to ignore complex Mayan suffixes.

However, our out-of-domain evaluation revealed strict limitations. Without the semantic grounding provided by authentic corpora, the model suffered from severe lexical hallucination when presented with out-of-vocabulary prompts. Furthermore, the model encountered a base-conversion bottleneck when attempting to reconcile decimal and vigesimal numerical systems. This highlights a distinct limitation of sequence-to-sequence arithmetic, as autoregressive transformers process numbers as semantic text tokens heavily biased by their base-10 pre-training, rather than as base-independent mathematical values. Finally, our ablation study demonstrated that employing a Multi-Task Learning architecture ultimately caused negative transfer, as auxiliary tasks competed for limited parameter capacity, exacerbating the tendency to overfit to synthetic templates.

Ultimately, we establish that synthetic data generation is a highly effective, ethically sound mechanism for building a grammatical baseline. However, the current evaluation methodology highlights that comparing synthetic training directly against out-of-domain organic testing introduces a severe domain mismatch. To isolate and resolve this semantic deficit, future work will transition to Phase 2 of our Curriculum Learning framework. We are currently implementing a Machine Translation Post-Editing (MTPE) pipeline with community volunteers, which will provide the scarce, high-quality organic data necessary to refine the model's lexical grounding.

\section{Ethical Considerations}
This research was conducted under a strict ``Zero Target-Language Scrape'' mandate to respect Indigenous data sovereignty. Publicly scraping parallel corpora frequently strips data of its cultural context and alienates the communities that produced it. All organic Q'eqchi' data utilized in this study was sourced from publicly available, community-curated dictionaries or provided directly by community partners with explicit permission for machine learning research. 

Furthermore, we recognize the inherent risk of structural bias when generating synthetic Indigenous data utilizing high-resource language priors. To mitigate algorithmic colonialism, our pipeline employs explicit debiasing mechanisms, and we establish that synthetic data must strictly serve as a structural primer, requiring authentic, community-driven data for final semantic refinement. Finally, by optimizing this pipeline for parameter-efficient fine-tuning on consumer-grade hardware and utilizing open-weights models, we ensure that the resulting codebases and adapters can be returned to, and independently run by, the communities they are designed to serve.

\begin{credits}
\subsubsection{\ackname} 
We extend our deepest gratitude to the volunteers of the Mayan Languages Preservation Project, the Project Director Dr. Winston K. Scott, and the community linguists who compiled the dictionaries and glossaries that made this research possible. This work was supported in part by the Undergraduate Research Program at the University of Houston. We also thank MasterWord Services, Inc. for providing industry context and supporting research into Indigenous language data sovereignty.

\subsubsection{\discintname}
The authors have no competing interests to declare that are relevant to the content of this article.
\end{credits}

\end{document}